\documentclass[runningheads]{llncs}

\usepackage{eccv}

\usepackage{eccvabbrv}

\usepackage{graphicx}
\usepackage{booktabs}
\usepackage{bm}

\usepackage[accsupp]{axessibility}  
\usepackage{lipsum}
\usepackage{subcaption}
\usepackage{floatrow}
\newfloatcommand{capbtabbox}{table}[][\FBwidth]
\usepackage{multirow}

\usepackage{hyperref}

\usepackage{orcidlink}

\begin{document}

\title{DiffuLT: How to Make Diffusion Model Useful for Long-tail Recognition} 


\author{Jie Shao \and Ke Zhu \and Hanxiao Zhang \and Jianxin Wu\thanks{J. Wu is the corresponding author.}}

\authorrunning{J.~Shao et al.}

\institute{State Key Laboratory for Novel Software Technology, Nanjing University, China \and School of Artificial Intelligence, Nanjing University, China
\email{\{shaoj,zhuk,zhanghx\}@lamda.nju.edu.cn}, \email{wujx2001@nju.edu.cn}}

\maketitle

\begin{abstract}  
  This paper proposes a new pipeline for long-tail (LT) recognition. Instead of re-weighting or re-sampling, we utilize the long-tailed dataset itself to generate a balanced proxy that can be  optimized through cross-entropy (CE). Specifically, a randomly initialized diffusion model, trained exclusively on the long-tailed dataset, is employed to synthesize new samples for underrepresented classes. Then, we utilize the inherent information in the original dataset to filter out harmful samples and keep the useful ones. Our strategy, \textbf{Diffu}sion model for \textbf{L}ong-\textbf{T}ail recognition (\textbf{DiffuLT}), represents a pioneering utilization of generative models in long-tail recognition. DiffuLT achieves state-of-the-art results on CIFAR10-LT, CIFAR100-LT, and ImageNet-LT, surpassing the best competitors with non-trivial margins. Abundant ablations make our pipeline interpretable, too. The whole generation pipeline is done without any external data or pre-trained model weights, making it highly generalizable to real-world long-tailed settings.
  
\end{abstract}

  \section{Introduction}

  Deep learning has exhibited remarkable success across a spectrum of computer vision tasks, especially in image classification~\cite{ResNet,VIT,SwinTransformer}. These models, however, encounter obstacles when faced with real-world long-tailed data, where the majority classes have abundant samples but the minority ones are sparsely represented. The intrinsic bias of deep learning architectures towards more populous classes exacerbates this issue, leading to sub-optimal recognition of minority classes despite their critical importance in practical applications.
  
  Conventional long-tailed learning strategies such as re-weighting~\cite{focal,ldam}, re-sampling~\cite{bbn,bags}, structural adjustments~\cite{ride,reslt}, share a commonality: they acknowledge dataset's imbalance and focus on the training of models. They demand meticulous design and are challenging to generalize. Recently, a shift towards studying the dataset itself and involving more training samples through external knowledge to mitigate long-tailed challenges has emerged. Examples include Mosaic~\cite{Mosaic} (which utilizes retrieval and filtering to augment object bounding boxes) and GIF~\cite{gif} (which expands limited datasets through synthetic sample generation via models like Stable Diffusion). Yet, in many real-world scenarios, access to specialized data is limited and closely guarded, such as in military or medical contexts. This predicament prompts a critical inquiry: \emph{Is it feasible to balance long-tailed datasets \emph{without} depending on external resources or models?}
  
  \begin{figure*}[tb]
    \centering
    \includegraphics[width=0.85\textwidth]{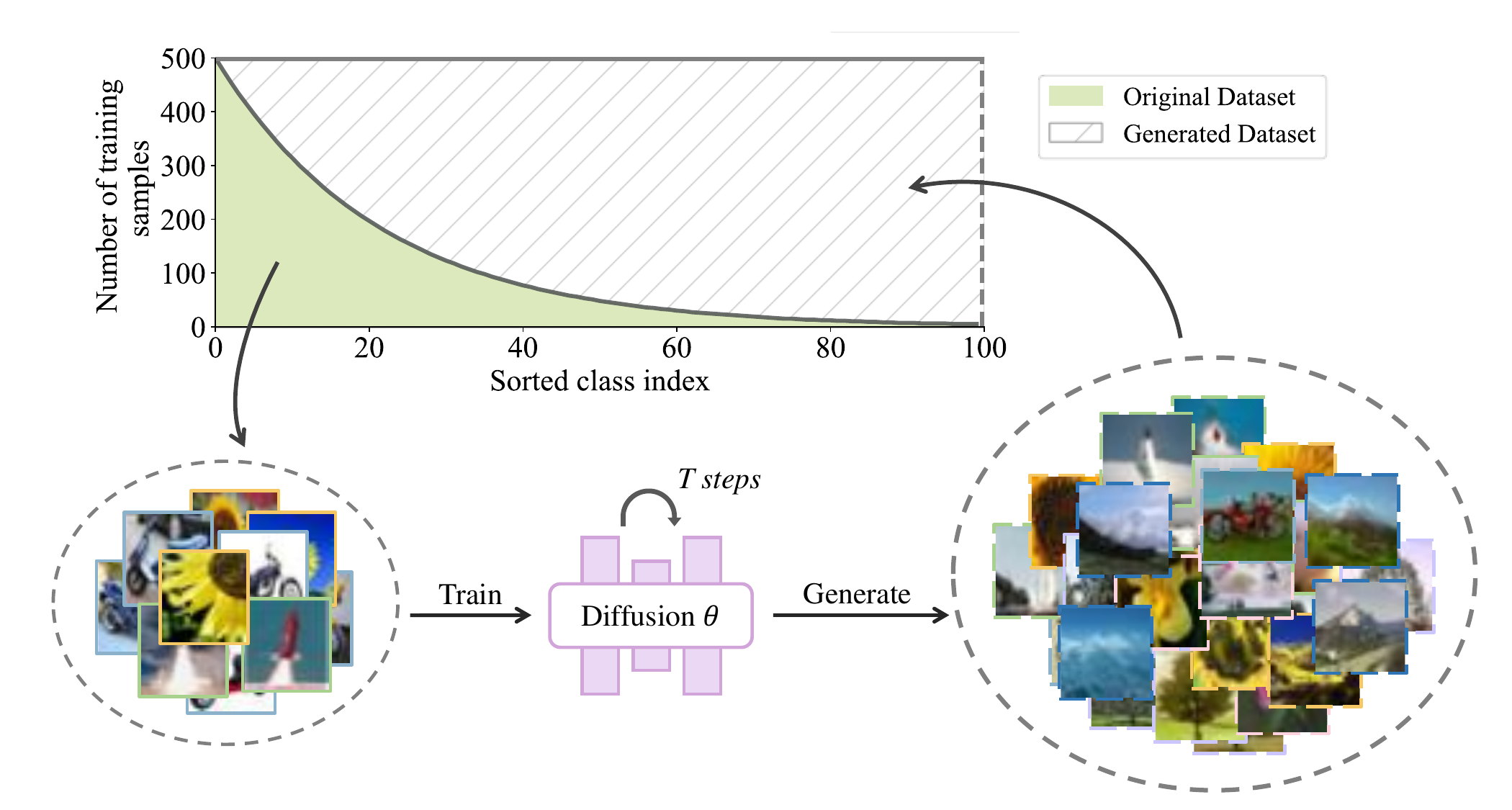}
    \caption{Our approach centers on the dataset, employing the long-tailed dataset to train a diffusion model, thereby creating a balanced proxy for direct optimization.}
    \label{fig:method}
  \end{figure*}
  
  The answer is yes. Recent diffusion models have demonstrated promising capability in producing high-fidelity images~\cite{ddpm,sde,ldm}. We thus propose a novel strategy as illustrated in \cref{fig:method}: training a diffusion model on a long-tailed dataset from scratch to generate new samples for tailed classes, thereby achieving dataset equilibrium. Subsequently, employing a simple network and training procedure (without specialized LT techniques), attains a balanced classifier. It is crucial to underscore the importance of training the diffusion model \emph{without external data or knowledge, to maintain fairness in comparison}.

  But, \cite{scaling} reveal that synthetic data are significantly less effective than real data in \emph{supervised} tasks. This limitation is especially evident for tail classes because it has too few data to train the diffusion model. We address this issue through a three-pronged strategy: enhancing the diffusion model's proficiency on tail classes, prioritizing the use of real data, and implementing a method to exclude detrimental synthetic data. We introduce an new pipeline, \textbf{DiffuLT} (\textbf{Diffu}sion model for \textbf{L}ong-\textbf{T}ail recognition), for long-tail datasets. It has four steps: initial training, sample generation, sample filtering, and retraining. Initially, a diffusion model is trained on the long-tailed dataset, incorporating a regularization term into the loss function to enhance its suitability for long-tail distributions. Subsequently, this model is employed to generate new samples, augmenting the dataset towards balance. Next, leveraging intrinsic dataset information, we filter out generated samples failing to surpass a predefined relevance threshold. The final step involves training a new classifier on this enriched dataset, with a minor adjustment to the loss function to reduce the impact of synthetic samples.
  
  Our comprehensive experimental analysis underscores the efficacy of our proposed method. In exploring the intricacies of our approach, we uncover intriguing phenomena. Notably, the diffusion model serves as a ``lecturer'' within the pipeline, aggregating dataset-wide information and distributing it across individual classes. Furthermore, we observe a strong correlation between the performance metrics of the diffusion model, specifically the Fréchet Inception Distance (FID), and the classification accuracy, highlighting the pivotal role of generative model quality in enhancing recognition performance.
  
  Our contributions are summarized as follows:
  \begin{itemize}
    \item We are the first to tackle long-tail recognition by synthesizing images \emph{without external data/models}. Then we simply use cross-entropy for classification. Our pipeline capitalizes on diffusion models specifically for long-tailed data, and solves the effectiveness problem of synthetic data.
    \item Through detailed analysis, we elucidate the pivotal role of the diffusion model within our pipeline and establish the correlation between the performance of the generative model and classification accuracy, offering valuable insights for future research endeavors.
    \item Extensive experimental validation across CIFAR10-LT, CIFAR100-LT, and ImageNet-LT datasets demonstrates the superior performance of our method over existing approaches.
  \end{itemize}

\section{Related Work}
\subsection{Long-tailed recognition} 
Long-tailed recognition is a challenging and practical task~\cite {LT_Effective_Number,LT_BBN,LT_LDAM,LTD_CAB}, since natural data often constitute a squeezed and imbalanced distribution. The majority of traditional long-tailed learning methods can be viewed as (or special cases) of re-weighting~\cite{LT_LDAM,LT_Decoupling,LT_Calibrate} and re-sampling~\cite{LT_Effective_Number}, with more emphasis on the deferred tail class to seek an optimization trade-off. There are variants of them that adopt self-supervised learning~\cite{MLS,LT_KD_SSL}, theoretical analysis~\cite{LT_logit_adjustgaussian,LT_logit_adjustment} and decoupling pipeline~\cite{LT_Decoupling,LT_BBN} to tackle long-tailed learning from various aspects, and they all achieve seemingly decent performance in downstream tasks.

One of the core difficulties in long-tailed learning is the \emph{in-sufficiency of tail samples}. And recently, quite an amount of works start to focus on this aspect by \emph{involving more training samples through external knowledge}. For example, Mosaic~\cite{Mosaic} and DLWL~\cite{dlwl} tries to adopt retrieval and filtering techniques to involve more object bounding boxes for long-tailed learning. By introducing more information, methods in this category often surpass traditional long-tailed methods and achieve state-of-the-art performance in recent pieces of literature~\cite{lpt,pel}.

Nevertheless, the most distinct drawback of previous works focusing on tail sample generation is that they either rely on \emph{external data source} or \emph{strong model weights}. This condition can seldomly hold true in practical scenarios where only a handful of \emph{specialized} data are available and are secretly kept (consider some important military or medical data). We thus raise a natural question about long-tailed learning: \emph{can we utilize the advantage of generating tail samples without resorting to any external data or model?} That is, the whole process is done in an in-domain (also called held-in) manner. In this paper, we propose to adopt the off-the-shelf diffusion model to learn and generate samples from the data at hand, and achieve quite satisfactory results.

\subsection{Diffusion models and synthetic data} 

Diffusion models have been highly competitive in recent years~\cite{ddpm,sde}. With promising image quality in both unconditional and conditional settings~\cite{guided_diffusion}, these models have rapidly advanced to produce high-resolution images with remarkable fidelity~\cite{ldm, dalle}. With the advent of products like Stable Diffusion and DALL·E, diffusion models have garnered significant attention across both academic and industrial landscapes. Research efforts have extended to exploring diffusion models in various scenarios, including few-shot~\cite{fsdm} and long-tailed generation~\cite{cbdm}. Despite the predominant use in creating digital art, the application of diffusion models in scenarios of limited data remains under-explored. This paper affirms the utility of diffusion models in enhancing representation learning, particularly within the long-tailed learning framework, offering a novel insight into their application beyond conventional generative tasks, and also makes those methods for generation under limited data~\cite{fsdm,cbdm} meaningful.

The integration of synthetic data into deep learning, generated through methods like GANs~\cite{gan,pix2pix} and diffusion models~\cite{guided_diffusion,ldm}, has been explored to enhance performance in image classification~\cite{active,synthetic,gif}, object detection~\cite{diffusionengine}, and semantic segmentation~\cite{datasetgan,diffusionengine}. These approaches often depend on substantial volumes of training data or leverage pre-trained models, such as Stable Diffusion, for high-quality data generation. Yet, the efficacy of generative models and synthetic data under the constraint of limited available data and in addressing imbalanced data distributions remains an unresolved inquiry. This paper specifically addresses this question, evaluating the utility of generative models and synthetic data in scenarios where data is scarce and imbalanced.

\section{Method}

\subsection{Preliminary}
For image classification, we have a long-tail dataset $\mathcal{D} = \{ (x_i, y_i) \}_{i=1}^N, y_i \in \mathcal{C}$ with each $x_i$ representing an input image and $y_i$ representing its corresponding label from the set of all classes $\mathcal{C}$. In the long-tail setting, a few classes dominate with many samples, while most classes have very few images, leading to a significant class imbalance. For each class $c_j$ in $\mathcal{C}$ ordered by sample count with $|c_1| \geq |c_2| \geq ... \geq |c_{M}|$, where $|c_j|$ denotes the number of samples in class $c_j$ and $|c_1| \gg |c_{M}|$, the ratio $r = \frac{|c_1|}{|c_{M}|}$ is defined as the long-tail ratio. The goal of long-tail classification is to learn a classifier $f_\varphi: \mathcal{X}  \rightarrow \mathcal{Y}$ capable of effectively handling the tail classes.

The naive idea is to train a generative model $\theta$ on the long-tail dataset $\mathcal{D}$ and use the trained model to generate new samples and supplement the tail classes. Currently, diffusion models are the most powerful generative models that excel at generating samples given a specified distribution. Due to their superior performance, we select diffusion models as the generative model in our pipeline. In our approach, we follow the Denoising Diffusion Probabilistic Model (DDPM~\cite{ddpm}) framework. Given a dataset $\mathcal{D} = \{ x_i, y_i \}_{i=1}^N$, we train a diffusion model to maximize the likelihood of the dataset. At every training step, we sample a mini-batch of images $\bm{x}_0$ from the dataset and add noise to obtain $\bm{x}_{t}$,
\begin{equation}
  q(\boldsymbol{x}_t \mid \boldsymbol{x}_0)=\mathcal{N}(\sqrt{\bar{\alpha}_t} \boldsymbol{x}_0, (1-\bar{\alpha}_t ) \boldsymbol{I})\,,
\end{equation}
where $\bar{\alpha}_t=\prod_{i=1}^t(1-\beta_i)$ is calculated through pre-defined variance schedule $\{\beta_t \in(0,1)\}_{t=1}^T$. After training a diffusion model $\theta$ to get $p_{\theta}(\bm{x}_{t-1} \mid \bm{x}_t, t)$, we reverse the above process step by step to recover the original image $\bm{x}_0$ from pure noise $\bm{x}_T \sim \mathcal{N}(\boldsymbol{0}, \boldsymbol{I})$, where the time step $T$ is the hyper-parameter to control the diffusion process's length. The training objective is to reduce the gap between the added noise in forward process and the estimated noise in reverse process. The loss function is defined as

\begin{equation}
  L_{\rm{DDPM}} =\mathbb{E}_{t \sim[1, T], \bm{x}_0, \epsilon_t} [ \|\boldsymbol{\epsilon}_t-\boldsymbol{\epsilon}_\theta (\sqrt{\bar{\alpha}_t} \bm{x}_0+\sqrt{1-\bar{\alpha}_t} \boldsymbol{\epsilon}_t, t ) \|^2 ]\,,
\end{equation}
where $\boldsymbol{\epsilon}_t \sim \mathcal{N}(\boldsymbol{0}, \boldsymbol{I})$ is the noise added in the forward process and $\boldsymbol{\epsilon}_\theta$ is the noise estimated by the trainable model with parameters $\theta$ in reverse process.

\subsection{Diffusion Model to Generate New Samples}
In this step, we train the \emph{randomly initialized} diffusion model $\theta$ to complement the dataset. Using a pre-trained generative model, such as Stable Diffusion trained on a large-scale dataset, for sample generation will definitely introduce information leakage and cause unfair comparison. So training a diffusion model from scratch becomes imperative. Here we make some modifications to the original DDPM to improve the controllability and quality of generated samples.

First, we make DDPM conditional, which means we can generate class-specific images based on the given label $y$. We incorporate the label $y$ directly as a conditional input, similar to time $t$, enabling the generation of class-specific images from the distribution $p_{\theta}(\bm{x}_{t-1} \mid \bm{x}_t, t, y)$ without adding new classifiers or structures. By transforming $y$ into a trainable class embedding integrated into the features of the model, we maintain the original loss function and make the model implicitly learn class information.

Second, we address the issue of class imbalance when applying the original DDPM to a long-tailed setting, where the model struggles to generate high-quality samples for less-represented (tail) classes. To counter this, we follow the idea of class-balancing diffusion models (CBDM~\cite{cbdm}). Specifically, we introduce a regularization term to the loss function. This term is designed to promote the generation of samples for tail classes. The loss function is defined as
\begin{equation}
  \begin{aligned}
    L_{\rm{CBDM}} = L_{\rm{DDPM}} &+\frac{\tau t}{|\mathcal{D}|} \sum_{y^{\prime} \in \mathcal{D}} ( \|\boldsymbol{\epsilon}_\theta (\boldsymbol{x}_t, y )-\operatorname{sg} (\boldsymbol{\epsilon}_\theta (\boldsymbol{x}_t, y^{\prime} ) ) \|^2\\
    &+\gamma \|\operatorname{sg} (\boldsymbol{\epsilon}_\theta (\boldsymbol{x}_t, y ) )-\boldsymbol{\epsilon}_\theta (\boldsymbol{x}_t, y^{\prime} ) \|^2 )\,,
  \end{aligned}
\end{equation}
where ``sg'' denotes stop gradient for the model, $\tau$, $\gamma$ are the hyper-parameters and $\mathcal{D}$ denotes the long-tailed dataset.

Upon training the model $\theta$ on long-tailed dataset $\mathcal{D}$, we use it to supplement the tail classes. We set a threshold $N_t$, and for any classes $c_j$ with fewer than $N_t$ samples, we generate $N_t - |c_j|$ images to reach the threshold. Specifically, for each class $c_j$, we configure the model's input class as $y = c_j$, and generate images by sampling from the conditional distribution $p_{\theta}(\bm{x}_{t-1} \mid \bm{x}_t, t, y)$. Initializing from pure Gaussian noise $\bm{x}_T$, we acquire images of class $c_j$ thereby. These generated images are straightforwardly labeled as $c_j$. 

This process results in a dataset of generated samples, $\mathcal{D}_{\rm{gen}} = \{ (x_i, y_i) \}_{i=1}^{N_{\rm{gen}}}$, where $N_{\rm{gen}} = \sum_{c_j \in \mathcal{C}} \max(0, N_t - |c_j|)$ is the total number of generated samples. Subsequently, training a classifier on the augmented dataset $\mathcal{D} \cup \mathcal{D}_{\rm{gen}}$ can significantly enhance performance on tail classes.

\subsection{Weighting and Filtering}
\label{section:filter}

Our experiments highlight the improvement of classifier performance brought by the generated samples in the last step (\cf \cref{tab:module}). However, two significant challenges emerge. First, the quality of generated samples cannot be assured to match that of the original ones. Second, while some generated samples contribute positively, others may detrimentally introduce noise to the classifier, undermining overall performance. To address these issues, it's essential not to treat all samples—--both original and generated—--equally. We adjust the importance of each sample and filter out those that tend to be harmful. This strategy acknowledges the varying contributions of different samples to the learning process, prioritizing original data and beneficial generated samples over others.

\textbf{Weighting} the generated samples is to decrease the importance of the generated samples and treat them as auxiliary data, which ensures the model prioritizes learning from the original dataset. To facilitate this, each image $x$ is assigned an additional label $y_g$, which distinguishes between generated and original samples. Specifically, $y_g = 1$ is used for generated samples, while $y_g = 0$ marks the original ones. During the training process, a weighted cross-entropy (WCE) loss function is employed. Considering the data pair $(x, y, y_g) \in \mathcal{D} \cup \mathcal{D_{\rm{gen}}} $, and the classifier $f_\varphi$ with parameters $\varphi$, the loss function is defined as
\begin{equation}
  L_{\rm{cls}}=-\sum\nolimits_{(x, y, y_g)\in \mathcal{D} \cup \mathcal{D_{\rm{gen}}}} (\omega y_g + (1 - y_g)) \log \frac{\exp (f_{\varphi,y}(x))}{\sum_{i=1}^M \exp (f_{\varphi, c_i}(x))}\,,
\end{equation}
where $\omega$ is the hyper-parameter to control the weight of generated samples, $f_{\varphi, y}(x)$ and $f_{\varphi, c_i}(x)$ are the logits of class $y$ and $c_i$ for image $x$, respectively. Given our premise that generated samples hold less significance than original data, we set $\omega$ within the [0, 1] interval.

\textbf{Filtering} is to remove the harmful generated samples. We leverage the existing information to judge the quality of generated images. So far, the original dataset $\mathcal{D}$ contains most of the information, thus the most direct way is to calculate the similarity between generated samples and original ones. For generated image $(x', y') \in \mathcal{D}_{\rm{gen}}$, we use L2 distance to find the closest images in $\mathcal{D}$ with the same label $y'$. We assume that any $x'$ that looks like any of the images in $\mathcal{D}$ with label $y'$ is a good sample. The filtering metric $d_1$ is defined as
\begin{equation}
  d_1(x', y') = \min\nolimits_{(x, y) \in \mathcal{D}, y = y'} \|x - x'\|\,.
\end{equation}
If $d_1(x', y')$ is greater than a set threshold $D_1$, we remove it. While the metric's basis is straightforward, it's uncertain if the most similar generated samples offer the greatest benefit to the classifier.

Given that instances tend to cluster more effectively in feature space than in pixel space, a viable approach involves projecting generated samples into feature space for filtration. To initiate this process, a classifier $f_0$ is trained on the original dataset $\mathcal{D}$, then we remove the linear head in $f_0$ to get the feature extractor $\overline{f}_0$. We define the filtering metric $d_2$ as
\begin{equation}
  d_2(x', y') = \min\nolimits_{(x, y) \in \mathcal{D}, y = y'} \|\overline{f}_0(x) - \overline{f}_0(x')\|\,.
\end{equation}
This metric implies that the well-generated samples should be close to the original samples in the feature space. The threshold $D_2$ is a hyper-parameter and samples with $d_2 < D_2$ are kept. The effectiveness of this metric hinges critically on the quality of the feature extractor.

Both metrics $d_1$ and $d_2$ share the inefficiency of requiring each generated image to be compared with all the original images from the same class, which is time-consuming. To address this, we introduce a new metric, $d_3$, which leverages the trained classifier $f_0$ without the need for direct comparison to the original dataset. We assume $f_0$ has assimilated the original dataset's information, enabling it to assess the quality of generated samples effectively. We define $d_3$ as the probability that $f_0$ assigns the label $y'$ to a generated sample $x'$:
\begin{equation}
  d_3(x', y') = \frac{\exp (f_{0,y'}(x'))}{\sum_{i=1}^M \exp (f_{0,c_i}(x'))}\,,
\end{equation}
If $d_3(x', y')$ is larger than a threshold $D_3$, we keep the generated sample. 

We introduce three filtering metrics, $d_1$, $d_2$, and $d_3$ to sift through the generated samples $\mathcal{D}_{\rm{gen}}$, discarding those that don't meet the requirement, resulting in a filtered dataset $\mathcal{D}_{\rm{filt}}$. The effectiveness and comparison of these metrics will be explored in the experiment section.

\subsection{Overall Pipeline}

\begin{figure*}[tb]
  \centering
  \includegraphics[width=0.9\textwidth]{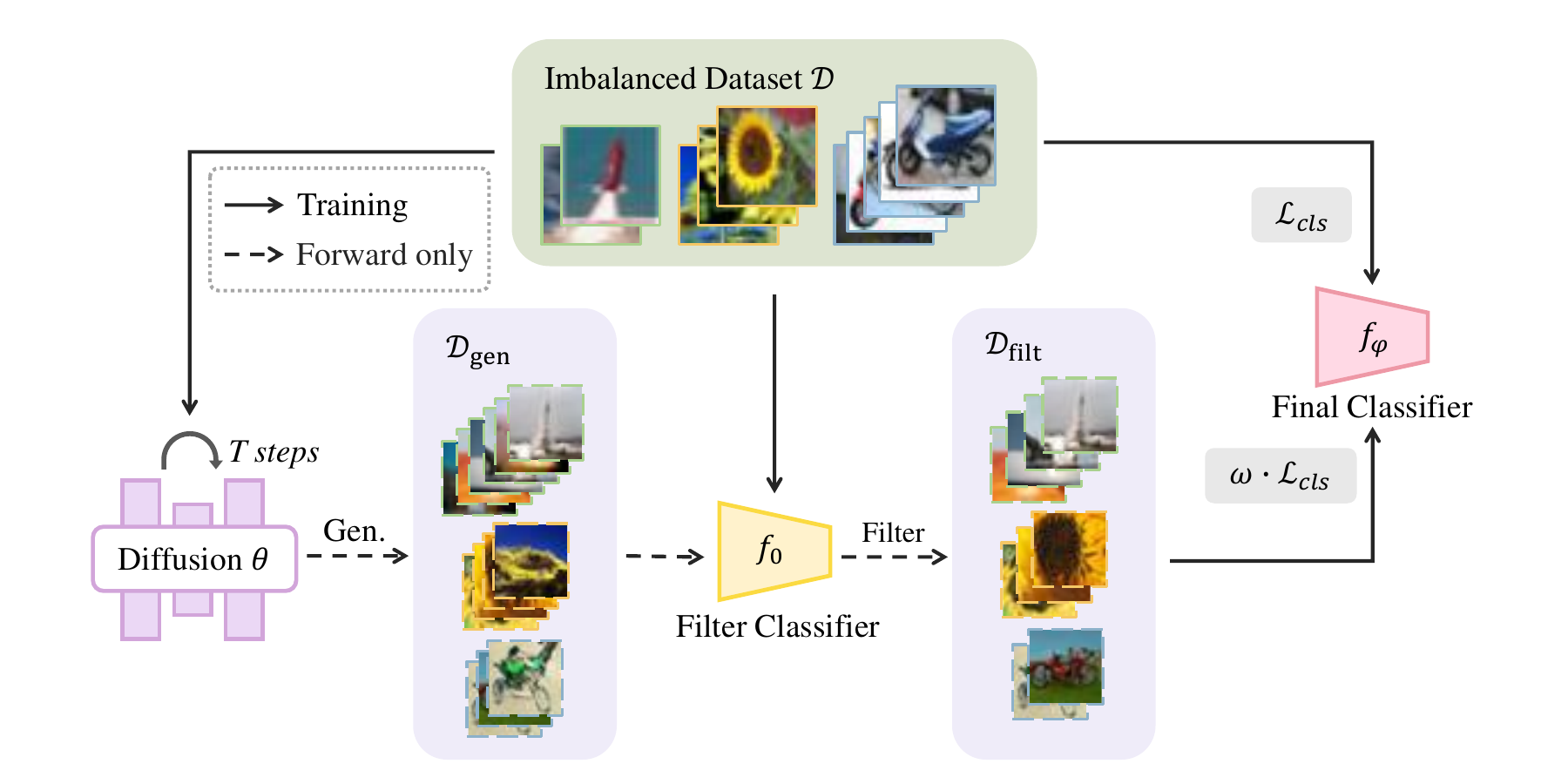}
  \caption{The overall pipeline of our method DiffuLT.}
  \label{fig:pipeline}
\end{figure*}

Now we propose a new pipeline called DiffuLT to address long-tail recognition. The pipeline is shown in \cref{fig:pipeline} with four steps:
\begin{itemize}
  \item \textbf{Training.} In the first step, we train a conditional, class-balanced diffusion model $\theta$ and a classifier $f_0$ using the original long-tailed dataset $\mathcal{D}$.
  \item \textbf{Generating.} We set the threshold $N_t$ and for each class with fewer than $N_t$ samples, we use the trained diffusion model $\theta$ to generate and supplement samples, resulting in a new dataset $\mathcal{D}_{\rm{gen}}$.
  \item \textbf{Filtering.} We check all generated samples, applying classifier $f_0$ or dataset $\mathcal{D}$ to compute the filtering metric ($d_1$, $d_2$ or $d_3$). Subsequently, we eliminate part of generated images to obtain the filtered dataset $\mathcal{D}_{\rm{filt}}$.
  \item \textbf{Training.} Finally we use the augmented dataset $\mathcal{D} \cup \mathcal{D}_{\rm{filt}}$ to train a new classifier $f_\varphi$ using weighted cross-entropy, serving as our final model.
\end{itemize}


\section{Experiment}
In this section, we show the experimental settings and results. We start with datasets and metrics. Then we proceed to benchmark our method against leading-edge approaches using CIFAR10-LT~\cite{ldam}, CIFAR100-LT~\cite{ldam}, and ImageNet-LT~\cite{ImageNetLT} datasets. Subsequent experiments are ablation studies conducted to identify the optimal settings of our method. Additionally, we offer insightful analyses to elucidate the diffusion model's impact within our methodology.

\subsection{Experimental setup}

\textbf{Datasets.} Our research evaluated three long-tailed datasets: CIFAR10-LT~\cite{ldam}, CIFAR100-LT~\cite{ldam}, and ImageNet-LT~\cite{ImageNetLT}, derived from the CIFAR10~\cite{CIFAR100},  CIFAR100~\cite{CIFAR100}, and ImageNet~\cite{ImageNet} datasets, respectively. The original CIFAR10 and CIFAR100 datasets comprise 50,000 images evenly distributed across 10 or 100 classes. Following the methodology described in~\cite{ldam}, we constructed long-tailed versions of these two datasets by adjusting the long-tail ratio $r$ to 100, 50, and 10 to test our method against various levels of class imbalance. ImageNet-LT, containing 115,846 images across 1,000 classes, was structured according to the guidelines provided in~\cite{ImageNetLT}, facilitating a comprehensive assessment of our approach under a wide spectrum of data imbalance scenarios.

\textbf{Baselines.} In our comparative analysis, we benchmark against a broad spectrum of classical and contemporary long-tailed learning strategies. For re-weighting and re-sampling techniques, we consider Cross-Entropy (CE), Focal Loss~\cite{focal}, LDAM-DRW~\cite{ldam}, cRT~\cite{crt}, BBN~\cite{bbn}, CSA~\cite{csa}, and ADRW~\cite{adrw}. Head-to-tail knowledge transfer approaches include M2m~\cite{m2m}, OLTR~\cite{oltr}, FSA~\cite{fsa}, and H2T~\cite{h2t}. Label-smoothing comparisons involve MisLAS~\cite{mislas}, DiVE~\cite{dive}, and SSD~\cite{ssd}, while for data-augmentation methodologies, we juxtapose CAM-BS~\cite{bags}, CMO~\cite{cmo}, and CUDA~\cite{cuda}. Lastly, SAM~\cite{sam} represents an advanced optimization technique in our evaluation.

\textbf{Implementation.} For CIFAR10/100-LT, we adopted the CBDM configuration, and set $\tau = 1$ and $\gamma = 0.25$. The generation thresholds $N_t$ for CIFAR10-LT and CIFAR100-LT were fixed at 5000 and 500, respectively. We employed ResNet-32 as the classifier backbone, applying a weighted cross-entropy loss with $\omega = 0.3$. According to \cref{tab:module} in our ablation study, the third filtering metric demonstrated superior performance, leading us to select a default threshold of $D_3 = 5\times 10 ^{-7}$. For ImageNet-LT experiments, we simplified the process by using guided diffusion~\cite{guided_diffusion} for generation without loss regularization or filtering, setting a generation threshold of $N_t=300$. The classifiers were based on ResNet-10 and ResNet-50 architectures, with the loss weight adjusted to $\omega = 0.5$. Those default settings were used in all experiments unless otherwise specified.

\subsection{Experimental Results}
This section presents a comparative analysis of our method against leading-edge techniques on CIFAR10-LT, CIFAR100-LT, and ImageNet-LT datasets. Specifically for CIFAR100-LT with an imbalanced ratio of 100 and ImageNet-LT, performance is assessed across four accuracy categories: overall, many-shot (classes with over 100 samples), medium-shot (classes with 20 to 100 samples), and few-shot (classes with fewer than 20 samples), providing a nuanced view of model effectiveness across varying class sample sizes.

\textbf{CIFAR100-LT \& CIFAR10-LT.} We benchmark our approach against a range of well-known and contemporary methods on the CIFAR100-LT and CIFAR10-LT datasets, with results detailed in \cref{tab:cifar}.  Methods like~\cite{BSCE,paco} utilize AutoAugment~\cite{autoaugment}, a data augmentation technique drawing on the \emph{external balanced} version of the dataset and significantly enhancing baseline performance (e.g., boosting CIFAR100-LT accuracy from 38.3\% to 45.3\% with $r=100$). To maintain fairness in evaluations, recent studies~\cite{unimix,reslt,sam,sbcl,adrw,csa,h2t} have excluded such results from their analyses. Aligning with this approach, we \textbf{limit our comparison to methods employing identical baselines as us and excluding external information}. On CIFAR100-LT, our method surpasses competing models, achieving accuracy improvements of 12.4\%, 12.0\%, and 7.9\% compared with the baseline for long-tail ratios $r=100$, 50, and 10, respectively, underscoring the significant efficacy of our approach. Similarly, on CIFAR10-LT, our model demonstrates strong competitiveness, enhancing accuracy by 13.9\%, 11.9\%, and 4.2\% across the long-tail ratios, further validating the robustness and effectiveness of our method.

For CIFAR100-LT with an imbalance ratio of $r=100$, we categorize performance into three groups based on class sample sizes: ``Many'' (classes with over 100 samples), ``Med.'' (classes with 20 to 100 samples), and ``Few'' (classes with fewer than 20 samples). The default settings for the methods in this analysis align with those in CIFAR100-LT's initial column, though deviations in experimental setups are indicated by $^*$. Specifically, CMO's reported statistics correspond to a model with an overall accuracy of 43.9\%, and CSA's with 45.8\%. While our approach does not lead in the ``Many'' category, it excels in ``Med.'' and ``Few'', significantly outperforming others in the ``Few'' group with a 29.0\% accuracy — 7.6\% above the nearest competitor and 19.9\% beyond the baseline. These findings underscore our method's distinct advantage in enhancing performance for tail classes, aligning with long-tail classification objectives and confirming its genuine effectiveness beyond merely favoring dominant classes.

\begin{table*}[tb]
	\small
	\centering
	\setlength{\tabcolsep}{5pt}
  \renewcommand{\arraystretch}{1.0}
  \caption{Results on CIFAR100-LT and CIFAR10-LT datasets. The imbalance ratio $r$ is set to 100, 50 and 10. The highest-performing results are highlighted in bold, with the second-best in underline. Additionally, we categorize CIFAR100-LT outcomes by class sample size - ``Many'', ``Med.'', and ``Few'' - at an imbalance ratio of $r=100$. It's noted that most models use consistent settings with their counterparts in CIFAR100-LT's primary analysis, with $^*$ indicating those under alternative configurations. We don't list methods that utilize external information here.}
	\begin{tabular}{l|ccc|ccc|ccc}
  \toprule
  \multirow{2}{*}{Method} & \multicolumn{3}{|c}{ CIFAR100-LT } & \multicolumn{3}{|c}{ CIFAR10-LT } & \multicolumn{3}{|c}{ Statistic (C100, 100) }\\
   & 100 & 50 & 10 & 100 & 50 & 10 & Many & Med. & Few\\
  \midrule CE & 38.3 & 43.9 & 55.7 & 70.4 & 74.8 & 86.4 & 65.2 & 37.1 & 9.1\\
  Focal Loss~\cite{focal} & 38.4 & 44.3 & 55.8 & 70.4 & 76.7 & 86.7 & 65.3 & 38.4 & 8.1 \\
  LDAM-DRW~\cite{ldam} & 42.0 & 46.6 & 58.7 & 77.0 & 81.0 & 88.2 & 61.5 & 41.7 & 20.2 \\
  cRT~\cite{crt} & 42.3 & 46.8 & 58.1 & 75.7 & 80.4 & 88.3 & 64.0 & 44.8 & 18.1 \\
  BBN~\cite{bbn} & 42.6 & 47.0 & 59.1 & 79.8 & 82.2 & 88.3 & - & - & - \\
  M2m~\cite{m2m} & 43.5 & - & 57.6 & 79.1 & -  & 87.5 & - & - & - \\
  CAM-BS~\cite{bags} & 41.7 & 46.0 & - & 75.4 & 81.4 & - & - & - & - \\
  MisLAS~\cite{mislas} & 47.0 & 52.3 & \underline{63.2} & 82.1 & \underline{85.7} & 90.0 & - & - & -  \\
  DiVE~\cite{dive} & 45.4 & 51.1 & 62.0 & - & - & - & - & - & - \\
  SSD~\cite{ssd} & 46.0 & 50.5 & 62.3 & - & - & - & - & - & -  \\
  CMO~\cite{cmo} & 47.2 & 51.7 & 58.4 & - & - & - & \textbf{70.4}$^*$ & 42.5$^*$ & 14.4$^*$\\
  SAM~\cite{sam} & 45.4 & - &	- & 81.9 & - & - & 64.4 & 46.2 & 20.8 \\
  CUDA~\cite{cuda} & 47.6 & 51.1 & 58.4 & - & - & - & 67.3 & \underline{50.4} & \underline{21.4} \\
  CSA~\cite{csa} & 46.6 & 51.9 & 62.6 & 82.5 & 86.0 & \textbf{90.8} & 64.3$^*$ & 49.7$^*$ & 18.2$^*$ \\
  ADRW~\cite{adrw} & 46.4 &	- &	61.9 & \underline{83.6} & - & 90.3 & - & - & -\\
  H2T~\cite{h2t} & \underline{48.9} & \underline{53.8} & - & - & - & - & - & - & -\\
  \midrule
  Diff * & \textbf{50.7} & \textbf{55.9} & \textbf{63.6} & \textbf{84.3} & \textbf{86.7} & \underline{90.6} & \underline{68.6} & \textbf{50.8} & \textbf{29.0}\\
  \bottomrule
  \end{tabular}
	\label{tab:cifar}
\end{table*}

\textbf{ImageNet-LT.} On the ImageNet-LT dataset, our methodology is evaluated against existing approaches, with findings summarized in \cref{tab:imagenet}. Utilizing a ResNet-10 backbone, our method registers a 49.9\% accuracy, outperforming the nearest competitor by 8.3\%. With ResNet-50, the accuracy further escalates to 56.2\%, marking a substantial 14.6\% enhancement over the baseline. Performance across class groupings—``Many'', ``Med.'', and ``Few''—mirrors the categorization strategy of CIFAR100-LT, offering insights into the method's class focus. Despite a slight decline in the ``Many'' category relative to the baseline, our approach excels in ``Med.'' and `Few'', with the latter witnessing a remarkable 33.4\% improvement over the baseline. These outcomes, consistent with CIFAR100-LT results, underscore our method's exceptional efficacy in boosting performance for tail classes, aligning perfectly with the objectives of long-tail classification.

\begin{table*}[tb]
  \setlength{\tabcolsep}{7pt}
  \renewcommand{\arraystretch}{1.0}
  \caption{Results on ImageNet-LT. We deploy ResNet-10 and ResNet-50 as classifier backbones. For the latter, we detail not only overall performance but also outcomes across distinct class sample size categories: ``Many'', ``Med.'', and ``Few''. Top-performing results are highlighted in bold, with second-best outcomes underlined.}
  \centering
  \footnotesize
  \begin{tabular}{l|c|cccc}
    \toprule & ResNet-10 & \multicolumn{4}{|c}{ ResNet-50 } \\
    & All & All & Many & Med. & Few \\
    \midrule CE & 34.8 & 41.6 & 64.0 & 33.8 & 5.8 \\
    Focal Loss~\cite{focal} & 30.5 & - & - & - & - \\
    OLTR~\cite{oltr} & 35.6 & - & - & - & - \\
    FSA~\cite{fsa} & 35.2 & - & - & - & - \\
    cRT~\cite{crt} & 41.8 & 47.3 & 58.8 & 44.0 & 26.1 \\
    BBN~\cite{bbn} & - & 48.3 & - & - & - \\
    MisLAS~\cite{mislas} & - & 52.7 & - & - & - \\
    CMO~\cite{cmo} & - & 49.1 & \textbf{67.0} & 42.3 & 20.5 \\
    SAM~\cite{sam} & & 53.1 & 62.0 & \underline{52.1} & 34.8 \\
    CUDA~\cite{cuda} & - & 51.4 & 63.1 & 48.0 & 31.1 \\
    CSA~\cite{csa} & \underline{42.7} & 49.1 & 62.5 & 46.6 & 24.1 \\
    ADRW~\cite{adrw} & - & \underline{54.1} & 62.9 & 52.6 & \underline{37.1}  \\
    \midrule
    Diff & \textbf{49.9} & \textbf{56.2} & \underline{63.2} & \textbf{55.4} & \textbf{39.2}  \\
    \bottomrule
    \end{tabular}
    \label{tab:imagenet}
\end{table*}

\subsection{Ablation Study}
This section presents ablation studies to evaluate the contribution of each component within our pipeline and optimize hyper-parameters. We standardize our experiments on the CIFAR100-LT dataset with an imbalance ratio of $r = 100$, employing ResNet-32 as the classifier backbone. Other settings are consistent with those in the main experiments unless otherwise specified.

\textbf{Different modules in our pipeline.} Our methodology is segmented into three principal components: the generative model (Gen), the weighted cross-entropy loss (Weight), and the filtering process (Filt), with the weighted loss using $\omega = 0.3$ and filtering employing the third metric $d_3$ at threshold $D_3 = 5\times 10^{-7}$, as determined from \cref{tab:filter}. We differentiate between the generative model following the original DDPM (Gen-D) and a variant incorporating a regularization term (Gen-C) to address the class imbalance. The baseline, devoid of these enhancements, sets the reference point. Results tabulated in \cref{tab:module}, reveal that even the basic DDPM integration boosts accuracy by 5.5\%. The CBDM model (Gen-C) further elevates performance by an additional 2.2\% over Gen-D. Both generative models benefit from the weighted loss, each seeing approximately a 3\% increase. Integrating Gen-C with both weighted loss and filtering culminates in a 50.7\% performance, evidencing the critical role each component plays in our pipeline's overall efficacy.

\textbf{Different weights in weighted cross-entropy loss.} In our exploration to optimize the weighted cross-entropy loss, we employed the CBDM generative model, excluding the filtering process, to assess the impact of varying weights ($\omega$). The results are summarized in \cref{tab:omega}. Treating generated samples on par with original ones ($\omega = 1.0$) yields a performance of 45.7\%. Iterative adjustments of $\omega$ demonstrate performance enhancements for any value between 0 and 1.0, with the optimal result of 48.4\% achieved at $\omega = 0.3$. These results affirm the utility of weighted cross-entropy loss in our framework, leading us to adopt $\omega = 0.3$ as the standard setting.

\begin{table*}[tb]
  \begin{floatrow}
  \setlength{\tabcolsep}{3pt}
  \capbtabbox{
  \renewcommand{\arraystretch}{1.0}
   \begin{tabular}{cccc|c}
   \toprule
    Gen-D & Gen-C & Weight & Filt & Acc. (\%) \\
   \midrule
     & & & & 38.3 \\
     \checkmark & & & & 43.8  \\
     & \checkmark & & & 45.6  \\
     \checkmark & & \checkmark & & 46.5 \\
     & \checkmark & \checkmark & & 48.4 \\
     & \checkmark & \checkmark & \checkmark & \textbf{50.7} \\
   \bottomrule
   \end{tabular}
  }{
   \caption{Results of different combinations of our proposed modules. Gen-D and Gen-C denote using the generative model following the DDPM and CBDM, respectively. Weight means using the weighted cross-entropy loss and Filt denotes using the filtering process.}
   \label{tab:module}
  }
  \setlength{\tabcolsep}{15pt}
  \capbtabbox{
	\renewcommand{\arraystretch}{0.9}
   \begin{tabular}{c|c}
   \toprule
   $\omega$ & Acc. (\%) \\
   \midrule
   0 & 38.3 \\
   0.1 & 46.2 \\
   0.3 & \textbf{48.4} \\
   0.5 & 48.1 \\
   0.7 & 47.3 \\
   0.9 & 46.6 \\
   1.0 & 45.6 \\
   \bottomrule
   \end{tabular}
  }{
   \caption{Performance of classifier with different weights $\omega$ in WCE. $\omega$ is in range of $[0, 1]$ and larger $\omega$ means the generated samples are more important during training.}
   \label{tab:omega}
  }
  \end{floatrow}
  \end{table*}

\textbf{Filtering metrics and thresholds.} In our experiments to identify the optimal filtering metric and its corresponding threshold, we utilized the CBDM generative model paired with a weighted cross-entropy loss set at $\omega = 0.3$. The methodology for metric calculation is elaborated in \cref{section:filter}. The table's first column introduces conditions under which generated samples are retained, based on each metric satisfying the stated inequality. The threshold $D$, a critical hyper-parameter for filtering, is determined through analysis of metric distributions within the training set, subsequently refining our search to the vicinity of each metric's peak performance as documented in \cref{tab:filter}. The table's baseline indicates the initial performance without filtering, with $\| \mathcal{D}_{\rm{filt}} \|$ representing the count of added samples.

Findings confirm the efficacy of all three filtering metrics in enhancing classifier performance, with the third metric, $d_3$, proving superior by achieving a peak accuracy of 50.7\%, a 2.3\% improvement over the baseline. Given the computational efficiency and simplicity, the third metric, $d_3$, is designated as our default filtering approach, with a set threshold of $D_3 = 5\times 10^{-7}$.

\begin{table*}[tb]
  \setlength{\tabcolsep}{10pt}
  \caption{Results of different filtering metrics and thresholds. The calculation of the filtering metrics $d_1$, $d_2$, and $d_3$ are shown in \cref{section:filter}. Generated samples that satisfy the rule are kept and the others are removed. $\| \mathcal{D}_{\rm{filt}} \|$ denotes the number of samples in the filtered dataset, i.e., the number of samples added to the original dataset.}
  \footnotesize
  \centering
  \renewcommand{\arraystretch}{0.95}
  \begin{tabular}{c|c|c|c}
    \toprule
    Metric \& Rule & Threshold $D$  & $\| \mathcal{D}_{\rm{filt}} \|$ & Acc. (\%)  \\
    \midrule
    w/o filt & - & 39153 & 48.4\\
    \midrule
    \multirow{3}{*}{$d_1 \leq D_1$} & 3000 & 18814 & 47.3 \\
    & 3500 & 29114 & \underline{49.7} \\
    & 4000 & 35274 & 48.4 \\
    \midrule
    \multirow{3}{*}{$d_2 \leq D_2$} & 8.5 & 25881 & 49.5 \\
    & 9 & 30504 & \underline{50.4} \\
    & 9.5 & 33727 & 48.7 \\
    \midrule
    \multirow{4}{*}{$d_3 \geq D_3$} & $5\times10^{-5}$ & 16475 & 49.0 \\
    & $5\times10^{-6}$ & 23498 & 50.0 \\
    & $5\times10^{-7}$ & 29223 & \underline{\textbf{50.7}} \\
    & $1\times10^{-7}$ & 32087 & 49.0 \\
    \bottomrule
    \end{tabular}
    \label{tab:filter}
\end{table*}

\subsection{Delving into DiffuLT}

\textbf{Role of diffusion model.} To decipher the efficacy and underlying mechanism of our pipeline, particularly the role of the diffusion model, we embarked on a series of investigative experiments. Utilizing original images from ``medium'' and ``few'' classes, alongside a variable proportion $p_{\rm{ma}}$ of ``many'' class images, we trained the diffusion model. This model then generaed samples exclusively for ``medium'' and ``few'' classes, with the collective performance of these classes denoted as $\rm{Acc}_{mf}$. Findings are consolidated in \cref{tab:role}, with the baseline scenario devoid of any generative process. The proportion $p_{\rm{ma}}$ spanned from 0\% to 100\%, where $p_{\rm{ma}} = 0\%$ implies reliance solely on ``medium'' and ``few'' class samples for diffusion model training, yielding a modest improvement of 1.0\% over the baseline to 26.0\%. Notably, as $p_{\rm{ma}}$ escalates, so does $\rm{Acc}_{mf}$, peaking at 32.8\% with $p_{\rm{ma}} = 100\%$, a 7.8\% enhancement compared to the baseline. This increment, achieved without generating ``many'' class samples, underscores the diffusion model's capacity to assimilate and relay information from populous to less represented classes. Such findings highlight the diffusion model's pivotal role in facilitating knowledge transfer across class groups, evidencing its indispensable contribution to our pipeline's success.

\textbf{FID vs Acc.} Our analysis, as presented in \cref{tab:module}, indicates that CBDM outperforms DDPM in classifier training, a finding further corroborated by CBDM's superior FID and IS metrics compared to DDPM. Intrigued by the potential correlation between generative model quality (FID/IS) and classifier performance, we experimented by varying the hyper-parameter $\tau$ within the regularization term of $L_{\rm{CBDM}}$ and observing the impact on classifier training with generated samples. The findings, detailed in \cref{tab:fid}, establish a clear positive relationship between the quality of the generative model (as indicated by FID and IS metrics) and classifier efficacy. Specifically, as FID improves and IS increases, classifier performance correspondingly elevates. This result gives us an insight that the diffusion model which can generate samples with a better quality is also superior in absorbing and combining the information from the original dataset altogether and transferring the information to the classifier

\begin{table*}[tb]
  \begin{floatrow}
  \setlength{\tabcolsep}{10pt}
  \capbtabbox{
	\renewcommand{\arraystretch}{0.85}
  \begin{tabular}{c|c}
    \toprule
    $p_{\rm{ma}}$ (\%) & $\rm{Acc}_{mf}$ (\%)  \\
    \midrule
    w/o gen & 25.0\\
    \midrule
    0  & 26.0 \\
    20  & 28.6 \\
    40 & 29.7 \\
    60 & 30.6 \\
    80 & 32.5 \\
    100  & \textbf{32.8}\\
    \bottomrule
    \end{tabular}
  }
  {
   \caption{Experiments aim to elucidate the source of performance enhancements within ``medium'' and ``few'' class groups and to clarify the diffusion model's role.}
   \label{tab:role}
  }
  \setlength{\tabcolsep}{10pt}
  \capbtabbox{
  \renewcommand{\arraystretch}{1.0}
    \begin{tabular}{l|ccc}
      \toprule
      $\tau$  & FID$\downarrow$ & IS$\uparrow$ & Acc. (\%) \\
      \midrule
      - & 7.17 & 13.29 & 46.6 \\
      1 & \textbf{5.37} & \textbf{13.32} & \textbf{48.4} \\
      2 & 6.82 &	11.58 &	46.0 \\
      3	& 10.41 & 10.11 & 44.8 \\
      4 & 13.30 & 9.34	& 43.9 \\
      5 & 19.51 & 8.43 & 43.3 \\
      \bottomrule
    \end{tabular}
  }
  {
   \caption{FID and IS of the generation model under different $\tau$ and its corresponding performance of the classifier. The first line denotes the original DDPM without any regularization term in loss.}
   \label{tab:fid}
  }
\end{floatrow}
\end{table*}

\section{Conclusion}
In this study, we adopt a novel data-centric approach to tackle the long-tail classification challenge. We introduce a pipeline that employs a diffusion model trained on the original dataset for sample generation, thereby enriching the dataset, followed by training a classifier on this augmented dataset. Additionally, we incorporate a weighted cross-entropy loss and a strategic filtering process to optimize the utilization of generated samples. Our method shows competitive performance, underlining its efficacy. The experiments highlight the critical role of the diffusion model and its performance within our framework. We posit that our approach represents a new avenue for addressing long-tail classification challenges, offering a valuable complement to existing strategies.

As for the limitations, it remains unclear how the final prediction model will behave if external data are involved in the diffusion training. We argue that if we train our generative (diffusion) model with abundant outer data sources, we might generate better synthetic samples, both for head and tail, thus making our final results even higher. We will leave this point as future work to further verify the generalized ability of our approach.

\clearpage
\bibliographystyle{splncs04}
\bibliography{main}
\end{document}